\documentclass[a4paper,conference]{IEEEtran}
\usepackage{blindtext, graphicx}
\usepackage{amsmath}
\usepackage{amssymb}
\usepackage{multirow}
\usepackage{url}
\usepackage{bbding}
\usepackage{pifont}
\usepackage{subfigure}
\usepackage{algorithm}
\usepackage{algorithmic}

\newcommand{\ie}{i.e.}
\newcommand{\etal}{\textit{et al.}}
\newcommand{\cmark}{\ding{51}}%
\newcommand{\xmark}{\ding{55}}%

\ifCLASSINFOpdf
\else
\fi
\begin{document}
%
\title{Affine Non-negative Collaborative Representation Based Pattern Classification}



%
\author{\IEEEauthorblockN{He-Feng Yin\IEEEauthorrefmark{1}\IEEEauthorrefmark{2},
Xiao-Jun Wu\IEEEauthorrefmark{1}\IEEEauthorrefmark{2},
Zhen-Hua Feng\IEEEauthorrefmark{3}\IEEEauthorrefmark{4} and
Josef Kittler\IEEEauthorrefmark{4}}
\IEEEauthorblockA{\IEEEauthorrefmark{1}School of Artificial Intelligence and Computer Science, Jiangnan University, Wuxi 214122, China\\ Email: 7141905017@vip.jiangnan.edu.cn, wu\_xiaojun@jiangnan.edu.cn}
\IEEEauthorblockA{\IEEEauthorrefmark{2}Jiangsu Provincial Engineering Laboratory of Pattern Recognition and Computational Intelligence\\
Jiangnan University, Wuxi 214122, China}
\IEEEauthorblockA{\IEEEauthorrefmark{3}Department of Computer Science, University of Surrey, Guildford GU2 7XH, UK\\ Email: z.feng@surrey.ac.uk}
\IEEEauthorblockA{\IEEEauthorrefmark{4}Centre for Vision, Speech and Signal Processing, University of Surrey, Guildford GU2 7XH, UK\\ Email: j.kittler@surrey.ac.uk}}


\maketitle

\begin{abstract}
During the past decade, representation-based classification methods have received considerable attention in pattern recognition. In particular, the recently proposed non-negative representation based classification (NRC) method has been reported to achieve promising results in a wide range of classification tasks. However, NRC has two major drawbacks. First, there is no regularization term in the formulation of NRC, which may result in unstable solution and misclassification. Second, NRC ignores the fact that data usually lies in a union of multiple affine subspaces, rather than linear subspaces in practical applications. To address the above issues, this paper presents an affine non-negative collaborative representation (ANCR) model for pattern classification. To be more specific, ANCR imposes a regularization term on the coding vector. Moreover, ANCR introduces an affine constraint to better represent the data from affine subspaces. The experimental results on several benchmarking datasets demonstrate the merits of the proposed ANCR method. The source code of our ANCR is publicly available at \url{https://github.com/yinhefeng/ANCR}.
\end{abstract}


%
\IEEEpeerreviewmaketitle

\section{Introduction}
Sparse representation-based classification remains one of the hot research topics in pattern recognition. The most two widely studied approaches include the classical sparse representation based classification (SRC)~\cite{wright2009robust} and collaborative representation based classification (CRC)~\cite{zhang2011sparse}. SRC directly employs all the training data as the dictionary and sparsely represents a test sample by solving the $\ell_1$-regularized minimization problem. Then the classification task is performed by checking which class yields the least reconstruction error (residual) of the test sample. SRC can achieve impressive classification performance in face recognition even when the test face image is occluded or corrupted. However, it is time consuming to solve the $\ell_1$-regularized minimization problem. To address this issue, Zhang \etal~\cite{zhang2011sparse} advocated the CRC method that uses the $\ell_2$-norm as the regularization term. 

Based on the classical SRC and CRC methods, researchers have proposed a variety of improved methods. Liu \etal~\cite{liu2016novel} presented a locally linear K nearest neighbor method that combines sparsity, locality, and reconstruction to obtain the reconstruction coefficients of a test sample. Song \etal~\cite{song2015progressive} proposed a progressive SRC algorithm using local discrete cosine transform for face classification. Lai \etal~\cite{lai2016classwise} developed a class-wise sparse representation method that seeks an optimum representation of the query image by minimizing the class-wise sparsity of the training data. Shao \etal~\cite{shao2017dynamic} proposed a new SRC-based face classification algorithm that exploits dynamic dictionary optimization on an extended dictionary using synthesized faces. Akhtar \etal~\cite{akhtar2017efficient} presented a sparsity augmented CRC (SA-CRC) scheme that augments a dense collaborative representation with a sparse representation. Inspired by SA-CRC, Li \etal~\cite{li2019sparsity} proposed a sparsity augmented weighted CRC approach for image recognition. Deng \etal~\cite{deng2018face} developed a superposed linear representation classifier to cast the recognition problem as one of representing the test image in terms of a superposition of the class centroids and shared intra-class differences. To address the problem of an insufficient number of training samples, Vo \etal~\cite{vo2018robust} developed a hierarchical CRC model. 
Song \etal~\cite{song2018dictionary} proposed the use of a 3D face model to synthesize faces with pose variations for pose-invariant face recognition. By employing only a subset of known pattern classes to collaboratively represent a test sample, Zheng \etal~\cite{zheng2019collaborative} designed a $k$ nearest classes-based CRC scheme. Xie \etal~\cite{xie2019sparse} presented an elastic-net regularized regression algorithm, which combines shared sparse representation with class specific CRC to represent a test sample. Song \etal~\cite{song2015half} proposed a half-face based CRC method to perform occlusion invariant face classification. Lan \etal~\cite{lan2020prior} proposed a  probabilistic CRC method for visual object recognition that includes the characteristics of the training samples of each class as prior knowledge. Gou \etal~\cite{gou2020weighted} developed a weighted discriminative CRC method that considers the competitive representation of each class and enhances the inter-class discrimination.

Xu \etal~\cite{xu2019sparse} pointed out that there exist negative elements in the coefficients obtained by SRC, CRC and their variants, which may result in misclassification of test samples. Motivated by non-negative matrix factorization (NMF)~\cite{lee1999learning}, they proposed a non-negative representation based classifier (NRC) that imposes a non-negative constraint on the coding vector. Extensive experiments on diverse classification tasks demonstrate the superiority of NRC over many existing representation based classification methods, including SRC, CRC and their variants.
Nevertheless, NRC has two shortcomings. First, due to the lack of regularization term on the coding vector, NRC may produce unstable solutions. Second, NRC cannot effectively deal with the data drawn from affine subspaces. To alleviate these drawbacks, we propose an affine non-negative collaborative representation (ANCR) model, in which a regularizer on the coding vector and the affine constraint are introduced.

\begin{figure}[!t]
\centering
\includegraphics[width=3in]{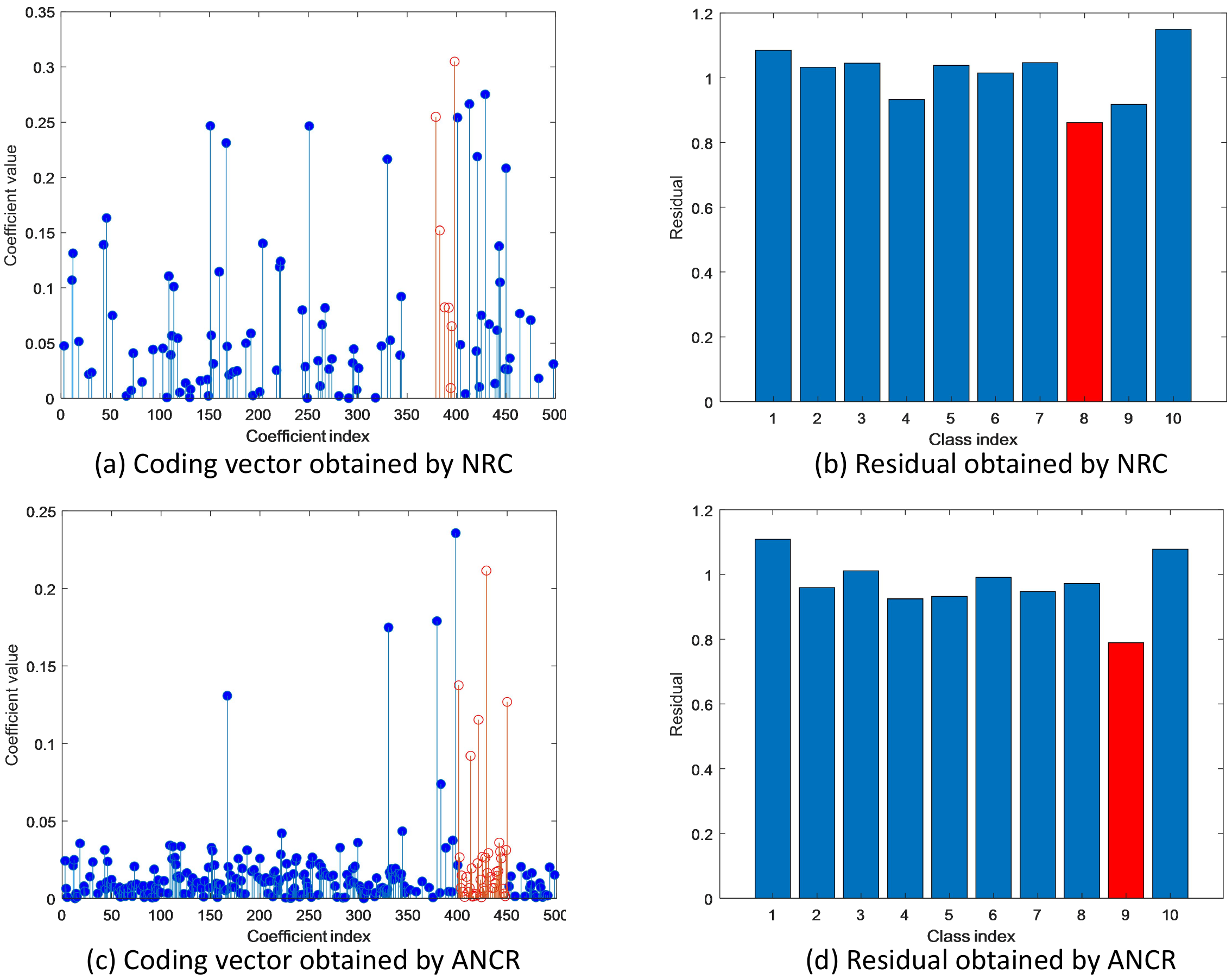}
\caption{Coding vectors and residuals obtained by NRC and ANCR: (a) coding vector obtained by NRC; (b) class-specific residual of NRC; (c) coding vector obtained by ANCR; (d) class-specific residual obtained by ANCR. The test sample belongs to the ninth class, i.e., digit 8. We can see that the eighth class yields the least residual for NRC, thus the test sample digit 8 is misclassified as digit 7. In contrast, we note that the ninth class 
has the minimum residual for our ANCR method, resulting in correct classification result.}
\label{fig:coeff_res}
\end{figure}

To illustrate the mechanism of ANCR, we conduct an experiment on the USPS dataset. This dataset contains images for digits 0-9. 50 images per class are used to form the training set. The training data matrix composed of the 500 images arranged in the order of $[0,1,2,\ldots,9]$, is denoted by $\mathbf{X}=[\mathbf{X}_0,\mathbf{X}_1,\ldots,\mathbf{X}_9]$. Consider a test sample from the ninth class (\ie, digit 8). The coding vector and residual obtained by NRC are shown in Fig.~\ref{fig:coeff_res} (a) and Fig.~\ref{fig:coeff_res} (b). The coding vector and residual obtained by ANCR are shown in Fig.~\ref{fig:coeff_res} (c) and Fig.~\ref{fig:coeff_res} (d). From Fig.~\ref{fig:coeff_res} (b), we can see that the eighth class has the least residual, \ie, the test sample is wrongly recognized as digit 7. Meanwhile, it can be seen in Fig.~\ref{fig:coeff_res} (a) that the eighth class has dominant coefficients. From Fig.~\ref{fig:coeff_res} (d), we note that the ninth class results in the minimum residual, \ie, the test sample is correctly recognized as digit 8. When we look more closely at the coding vectors of NRC and ANCR for the eighth class (indices 351-400), the number of nonzero entries are 7 and 14, respectively. In contrast, for the ninth class (indices 401-450), the number of nonzero entries of NRC and ANCR are 16 and 41, respectively. By introducing the coding vector regularizer and affine constraint into NRC, more coefficients are concentrated on the correct class, resulting in improved performance. 

The main contributions of the paper are as follows, 
\begin{itemize}
\item We propose an affine non-negative collaborative representation (ANCR) model by regularizing the coding vector and introducing an affine constraint in the formulation of NRC.
\item An optimisation algorithm based on the alternating direction method of multipliers (ADMM)~\cite{boyd2011distributed} is developed to efficiently solve the optimization problem of ANCR.
\item An extensive evaluation of the proposed ANCR method on diverse benchmarking datasets demonstrate that it outperforms conventional representation based classification methods as well as some deep learning based methods.
\end{itemize}

\section{Related Work}
\label{Sec_2}
In recent years, the non-negative constraint has attracted considerable attention in various research areas. For instance, Vo \etal~\cite{vo2009nonnegative} developed a non-negative least square (NNLS) algorithm for face recognition, which represents each new sample as a non-negative linear combination of training samples. Zhuang \etal~\cite{zhuang2012non} proposed a non-negative low-rank and sparse (NNLRS) graph for semi-supervised learning. Yin \etal~\cite{yin2015laplacian} presented a non-negative sparse hyper-Laplacian regularized LRR model (NSHLRR) for unsupervised and semi-supervised applications. Xu \etal~\cite{xu2018non} developed a unified non-negative subspace representation constrained leaning scheme for discriminative correlation filter (DCF) based tracking. Zhang \etal~\cite{zhang2018normalized} presented a normalized non-negative sparse coding ($\textrm{N}^3\textrm{SC}$) method that enforces both the non-negative constraint and the shift-invariant constraint to the traditional sparse coding criteria. Chen \etal~\cite{chen2019non} designed a non-negative representation based discriminative dictionary learning algorithm (NRDL) for face classification. Xu \etal~\cite{xu2019scaled} proposed a scaled simplex representation (SSR) for subspace clustering. Xu \etal~\cite{xu2019non} designed a jointly non-negative, sparse and collaborative representation (NSCR) for image recognition. Zhao \etal~\cite{zhao2020laplacian} developed a Laplacian regularized non-negative representation (LapNR) method for clustering and dimensionality reduction tasks. Inspired by NRC, Yin \etal~\cite{yin2020class} explored a class-specific residual constraint non-negative representation (CRNR) for pattern classification.

The affine constraint has also been widely exploited in clustering and classification tasks. In sparse subspace clustering (SSC), Elhamifar \etal~\cite{elhamifar2009sparse,elhamifar2013sparse} introduced the affine constraint to deal with affine subspaces. Xu \etal~\cite{xu2017affine} proposed an affine subspace clustering (ASC) algorithm which incorporates the affine constraint into a ridge regression formulation of the sparse representation problem. Dong \etal~\cite{dong2019robust} presented a sparse representation based affine subspace clustering method, which employs the non-convex smoothed $\ell_0$-norm to replace the $\ell_0$ norm. In locality-constrained linear coding (LLC)~\cite{wang2010locality}, the affine constraint is also referred to as the shift-invariant constraint. Motivated by LLC, Wei \etal~\cite{wei2013locality} developed a locality-sensitive dictionary learning algorithm for SRC, in which the designed dictionary is capable of preserving the local data structure. Benuma \etal~\cite{benuwa2020kernel} proposed a kernel locality sensitive discriminative sparse representation (K-LSDSR) for face recognition.

\section{The Proposed ANCR Method}
\label{Sec_3}
In this section, we first present a mathematical formulation of our ANCR approach. Then we introduce a detailed solution of the learning ANCR problem. Last, we provide an ANCR-based classification method. For the ease of presentation, we introduce the notation used in this paper. Assume that we have $n$ training samples belonging to $K$ classes, and let the training data matrix be denoted by $\mathbf{X}=\left[\mathbf{X}_{1}, \mathbf{X}_{2}, \ldots, \mathbf{X}_{K}\right] \in \mathbb{R}^{d \times n}$, where $\mathbf{X}_i$ is the data matrix of the $i$-th class. The $i$-th class has $n_i$ training samples and $\sum_{i=1}^Kn_i=n$ ($i=1,2,\ldots,K$), $d$ is the dimensionality of the vectorized samples.

\subsection{Formulation of ANCR}
To overcome the drawbacks of NRC, we incorporate the elastic-net regularizer~\cite{zou2005regularization}, \ie, $\lambda_1\left \| \boldsymbol{c} \right \|_2^2+\lambda_2\left \| \boldsymbol{c} \right \|_1$, into the formulation of NRC. The elastic-net regularizer is a combination of the $\ell_1$-norm and $\ell_2$-norm. Therefore, it has the merits of both the lasso and ridge regression methods. We introduce the affine constraint and obtain the following optimization problem,
\begin{equation}
\label{obj1_ancr}
\begin{split}
&\underset{\boldsymbol{c}}{\textrm{min}} \ \left \| \boldsymbol{y}-\mathbf{X}\boldsymbol{c} \right \|_2^2+\lambda_1\left \| \boldsymbol{c} \right \|_2^2+\lambda_2\left \| \boldsymbol{c} \right \|_1. \\ & \textrm{s.t.} \ \boldsymbol{c}\geq 0,\boldsymbol{1}^T\boldsymbol{c}=1
\end{split}
\end{equation}
It is worth noting that since $\boldsymbol{c}\geq 0$ and $\left \| \boldsymbol{c} \right \|_1=\boldsymbol{1}^T\boldsymbol{c}=1$. By removing the $\ell_1$-norm constraint, Eq. (\ref{obj1_ancr}) can be reformulated as,
\begin{equation}
\label{obj_ancr}
\underset{\boldsymbol{c}}{\textrm{min}} \ \left \| \boldsymbol{y}-\mathbf{X}\boldsymbol{c} \right \|_2^2+\lambda\left \| \boldsymbol{c} \right \|_2^2, \ \textrm{s.t.} \ \boldsymbol{c}\geq 0,\boldsymbol{1}^T\boldsymbol{c}=1
\end{equation}
The above equation is the objective function of the proposed ANCR method.

\subsection{Optimization}
We adopt an alternative optimization strategy to solve the ANCR problem. By introducing an auxiliary variable $\boldsymbol{z}$, Eq.~(\ref{obj_ancr}) can be rewritten as,
\begin{equation}
\label{equi_prob}
\underset{\boldsymbol{c},\boldsymbol{z}}{\textrm{min}} \ \left \| \boldsymbol{y}-\mathbf{X}\boldsymbol{c} \right \|_2^2+\lambda\left \| \boldsymbol{c} \right \|_2^2, \ \textrm{s.t.} \ \boldsymbol{z}=\boldsymbol{c},\boldsymbol{z}\geq0,\boldsymbol{1}^T\boldsymbol{z}=1
\end{equation}

The above optimization problem can be solved using the ADMM~\cite{boyd2011distributed} method. Specifically, the Lagrangian function of Eq.~(\ref{equi_prob}) is,
\begin{equation}
\label{lagrange}
\begin{split}
\mathcal{L} (\boldsymbol{c},\boldsymbol{z},\boldsymbol{\delta},\rho) =&\left \| \boldsymbol{y}-\mathbf{X}\boldsymbol{c} \right \|_2^2+\lambda\left \| \boldsymbol{c} \right \|_2^2+<\boldsymbol{\delta},\boldsymbol{z}-\boldsymbol{c}>\\&+\frac{\rho}{2}\left \| \boldsymbol{z}-\boldsymbol{c} \right \|_2^2,
\end{split}
\end{equation}
where $\boldsymbol{\delta}$ is the Lagrange multiplier and $\rho>0$ is a penalty parameter. The optimization of Eq.~(\ref{lagrange}) can be solved iteratively by updating $\boldsymbol{c}$ and $\boldsymbol{z}$ one at a time. The detailed updating procedures are presented as follows.

\textit{Update} $\boldsymbol{c}$: Fix the other variables and update $\boldsymbol{c}$ by solving the following problem,
\begin{equation}
\label{prob_c}
\underset{\boldsymbol{c}}{\textrm{min}} \ \left \| \boldsymbol{y}-\mathbf{X}\boldsymbol{c} \right \|_2^2+\lambda\left \| \boldsymbol{c} \right \|_2^2+<\boldsymbol{\delta},\boldsymbol{z}-\boldsymbol{c}>+\frac{\rho}{2}\left \| \boldsymbol{z}-\boldsymbol{c} \right \|_2^2
\end{equation}

Setting the partial derivative of Eq.~(\ref{prob_c}) with respect to $\boldsymbol{c}$ to zero, we can obtain the following closed-form solution of $\boldsymbol{c}$,
\begin{equation}
\label{solu_c}
\boldsymbol{c}_{t+1}=[\mathbf{X}^T\mathbf{X}+\frac{(\rho+2\lambda)}{2}\mathbf{I}]^{-1}[\mathbf{X}^T\boldsymbol{y}+\frac{\rho \boldsymbol{z}_t+\boldsymbol{\delta}_t}{2}]
\end{equation}

\textit{Update} $\boldsymbol{z}$: To update $\boldsymbol{z}$, we fix other variables and solve the following problem,
\begin{equation}
\label{prob_z}
\underset{\boldsymbol{z}}{\textrm{min}} \ \left \| \boldsymbol{z}-(\boldsymbol{c} -\frac{\boldsymbol{\delta}}{\rho})\right \|_2^2, \ \textrm{s.t.}  \ \boldsymbol{z}\geq0,\boldsymbol{1}^T\boldsymbol{z}=1
\end{equation}
The above problem can be solved by the algorithm proposed by Huang \etal~\cite{huang2015new}.

\textit{Update} $\boldsymbol{\delta}$: 
The Lagrange multiplier $\boldsymbol{\delta}$ is updated as
\begin{equation}
\label{update_delta}
\boldsymbol{\delta}_{t+1}=\boldsymbol{\delta}_t+\rho(\boldsymbol{z}_{t+1}-\boldsymbol{c}_{t+1})
\end{equation}

The above procedures for  solving Eq.~(\ref{obj_ancr}) are summarized in Algorithm~\ref{alg1}.
\begin{algorithm}[!t]
\caption{Solve Eq.~(\ref{obj_ancr}) via ADMM} 
\label{alg1} 
\begin{algorithmic}[1]
\REQUIRE Test sample $\boldsymbol{y}$, training data matrix $\mathbf{X}$, balancing parameter $\lambda$, $\textrm{tol}>0$, $\rho>0$ and the maximum iteration number $T$.
\STATE Initialize $\boldsymbol{z}_0=\boldsymbol{c}_0=\boldsymbol{\delta}_0=\boldsymbol{0}$;
\WHILE{not converged} 
\STATE Update $\boldsymbol{c}$ by Eq.~(\ref{solu_c});
\STATE Update $\boldsymbol{z}$ by solving Eq.~(\ref{prob_z});
\STATE Update $\boldsymbol{\delta}$ by Eq.~(\ref{update_delta});
\ENDWHILE 
\ENSURE Coding vectors $\boldsymbol{z}$ and $\boldsymbol{c}$.
\end{algorithmic} 
\end{algorithm}

\subsection{Classification}
Given a test sample $\boldsymbol{y}\in\mathbb{R}^d$, we first obtain its coding vector $\boldsymbol{c}$ over the entire training data $\mathbf{X}$ by Algorithm~\ref{alg1}. Then the test sample is assigned into the class that yields the least residual, \ie, $\textrm{identity}(\boldsymbol{y})=\textrm{arg} \ \underset{i}{\textrm{min}}\left \| \boldsymbol{y}-\mathbf{X}_i\boldsymbol{c}_i \right \|_2$, where $\boldsymbol{c}_i$ is the coding vector that belongs to the $i$-th class. The complete process of the proposed ANCR method is summarized in Algorithm~\ref{alg2}.
\begin{algorithm}[t]
\begin{algorithmic}[1]
\vspace{0.03in}
\REQUIRE Training data matrix $\mathbf{X}=\left[\mathbf{X}_{1}, \mathbf{X}_{2}, \ldots, \mathbf{X}_{K}\right] \in \mathbb{R}^{d \times n}$, test data $\boldsymbol{y} \in \mathbb{R}^{d}$ and balancing parameter $\lambda$.
\STATE Normalize the columns of $\mathbf{X}$ and $\boldsymbol{y}$ to have unit $\ell_2$ norm;
\STATE Obtain the coding vector $\boldsymbol{c}$ of $\boldsymbol{y}$ on $\mathbf{X}$ by solving the ANCR model in Eq.~(\ref{obj_ancr});
\STATE Compute the class-specific residuals $\boldsymbol{r}_i=\left \| \boldsymbol{y}-\mathbf{X}_i\boldsymbol{c}_i \right \|_2$;
\ENSURE $\textrm{label}(\boldsymbol{y})=\arg \underset{i}{\textrm{min}}\left(\boldsymbol{r}_{i}\right)$
\vspace{0.03in}
\end{algorithmic}
\caption{Our proposed ANCR algorithm}
\label{alg2}
\end{algorithm}

\section{Experiments and Analysis}
\label{Sec_4}
In this section, we evaluate the classification performance of ANCR on several benchmarking datsets, including the AR database~\cite{martinez1998ar} for face recognition, USPS dataset~\cite{hull1994database} for handwritten digit classification, Stanford 40 Actions dataset~\cite{yao2011human} for action recognition, three fine-grained object datasets which includes the CUB-200-2011 dataset~\cite{WahCUB_200_2011}, the Aircraft dataset~\cite{maji2013fine} and the Cars dataset~\cite{krause20133d}. We compare the classification accuracy of ANCR with NSC~\cite{lee2005acquiring}, linear SVM, SRC~\cite{wright2009robust}, CRC~\cite{zhang2011sparse}, CROC~\cite{chi2014classification}, ProCRC~\cite{cai2016probabilistic}, SA-CRC~\cite{akhtar2017efficient}, CCRC~\cite{yuan2018a} and NRC~\cite{xu2019sparse}. In addition, on the Aircraft and Cars datasets, we also compare ANCR with key deep methods, such as VGG19~\cite{simonyan2015very}, FV-FGC~\cite{gosselin2014revisiting}, and B-CNN~\cite{lin2015bilinear}.

\subsection{Experiments on the AR Database}
The AR database~\cite{martinez1998ar} comprises over 4000 frontal images for 126 individuals, these images contain variations in facial expressions, illumination and occlusions. Following the experimental settings in~\cite{xu2019sparse}, in our experiments, we use the subset with only illumination and expression changes. This subset contains 50 male and 50 female subjects. For each subject, 7 images from Session 1 are used as training samples, and 7 images from Session 2 are used as test samples. All the images are cropped to 60$\times$43 pixels and projected to a subspace of dimensions 54, 120, and 300 by PCA. The experimental results are listed in Table~\ref{tab:tab_ar}. Note, the balancing parameter $\lambda$ of ANCR is set to $1e-4$. We can observe that our proposed ANCR consistently outperforms the other methods under all the three reduced dimensions.
\begin{table}[!t]
\renewcommand{\arraystretch}{1.3}
\caption{A comparison of the proposed method in terms of recognition accuracy (\%) on the AR database.}
\label{tab:tab_ar}
\centering
\begin{tabular}{lccc}
\hline
Dimensions     & 54  & 120        & 300        \\ \hline
NSC~\cite{lee2005acquiring}  & 70.7    & 75.5   & 76.1   \\
SVM   & 81.6     & 89.3    & 91.6   \\
SRC~\cite{wright2009robust}  & 82.1   & 88.3    & 90.3   \\
CRC~\cite{zhang2011sparse} & 80.3   & 90.1      & 93.8 \\
CROC~\cite{chi2014classification}  & 82.0      & 90.8     & 93.7  \\
ProCRC~\cite{cai2016probabilistic} & 81.4     & 90.7     & 93.7  \\
SA-CRC~\cite{akhtar2017efficient} &   75.1   &   86.7   &  92.5 \\
CCRC~\cite{yuan2018a} &   81.2   &  88.9    &  93.1 \\
NRC~\cite{xu2019sparse}  & 85.2     & \textbf{91.3}     & 93.3  \\
ANCR &  \textbf{86.0}   &  \textbf{91.3}  &  \textbf{94.0}  \\
\hline
\end{tabular}
\end{table}

\subsection{Experiments on the USPS Dataset}
The USPS dataset~\cite{hull1994database} consists of 9298 images of digit numbers, i.e. from 0 to 9. The training set and test set of USPS have 7291 and 2007 images, respectively. All the images are resized into 16$\times$16 pixels. For each class, $N$ ($N$=50, 100, 200, 300) images from the training set are randomly selected for training and all the images in the test set are used for testing. Experiments are repeated for 10 times and the average results are reported in Table~\ref{tab:tab_usps}. The balancing parameter $\lambda$ of ANCR is set to 0.001. We can see that ANCR achieves the best recognition result for all the settings. With the increase of the number of training images, the recognition accuracy of all the approaches improves steadily.

\begin{table}[!t]
\renewcommand{\arraystretch}{1.3}
\caption{A comparison of the proposed method in terms of recognition accuracy (\%) on the USPS dataset.}
\label{tab:tab_usps}
\centering
\begin{tabular}{lcccc}
\hline
$N$     & 50  & 100    & 200     & 300        \\ \hline
NSC~\cite{lee2005acquiring}     & 91.2             & 92.2   & 92.8  & 92.8 \\
SVM   & 91.6             & 92.5    & 93.1 & 93.2  \\
SRC~\cite{wright2009robust}    &       89.1      &  91.2   & 92.9  & 93.8  \\
CRC~\cite{zhang2011sparse}    & 89.8             & 90.8      & 91.5 & 91.5\\
CROC~\cite{chi2014classification} & 91.9      & 91.3     & 91.7 & 91.8 \\
ProCRC~\cite{cai2016probabilistic}    & 90.9     & 91.9     & 92.2 & 92.2 \\
SA-CRC~\cite{akhtar2017efficient}  &   86.4   &   88.5   &  90.5  &  91.3 \\
CCRC~\cite{yuan2018a}  &  90.9    &  92.2   &  93.1   &  93.0 \\
NRC~\cite{xu2019sparse}  & 90.3        & 91.6     & 92.7 & 93.0 \\
ANCR  &  \textbf{92.1}   & \textbf{93.0}   & \textbf{93.5} &  \textbf{94.3} \\
\hline
\end{tabular}
\end{table}

\subsection{Experiments on the Stanford 40 Actions Dataset}
\begin{figure}[!t]
\centering
\includegraphics[width=3in]{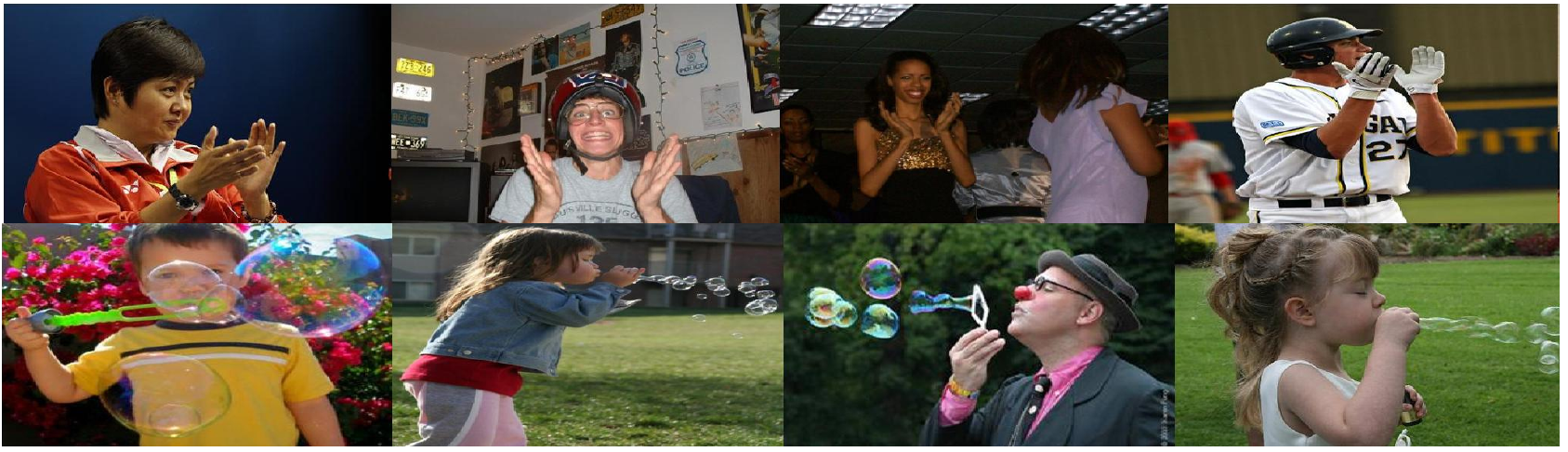}
\caption{Example images from the Stanford 40 Actions dataset.}
\label{fig:exam_St40}
\end{figure}

The Stanford 40 Actions dataset~\cite{yao2011human} has 40 different classes of human actions, e.g., applauding, blowing bubbles, and cooking. Some example images from the dataset are shown in Fig.~\ref{fig:exam_St40}. This dataset contains 9352 images in total, 180 to 300 images per action. Following the training-test split settings scheme in~\cite{xu2019sparse}, we randomly select 100 images per class as the training images and use the remaining images as the test set. The balancing parameter $\lambda$ of ANCR is set to 0.001. We extract image features by the pre-trained VGG-19 network. The dimensionality of the extracted VGG feature vector of each image is 4096. AlexNet~\cite{krizhevsky2012imagenet}, EPM~\cite{sharma2013expanded}, ASPD~\cite{khan2015recognizing} and VGG-19~\cite{simonyan2015very} are used for comparison. EPM and ASPD are two leading approaches for action recognition with still images. Experimental results are reported in Table~\ref{tab:tab_st40}. One can see that ANCR outperforms both the conventional representation based classification methods and the deep learning models in terms of recognition accuracy, which again validates the superiority of the proposed method.

\begin{table}[!t]
\renewcommand{\arraystretch}{1.3}
\caption{A comparison of the proposed method in terms of recognition accuracy (\%) on the Stanford 40 Actions dataset.}
\label{tab:tab_st40}
\centering
\begin{tabular}{lc|cc}
\hline
Methods     & Accuracy & Methods        & Accuracy        \\ \hline
Softmax  &   77.2  &  CCRC~\cite{yuan2018a}  & 79.2   \\
NSC~\cite{lee2005acquiring}  &   74.7  &   NRC~\cite{xu2019sparse}  &  81.1  \\
SRC~\cite{wright2009robust}  & 78.7   &  AlexNet~\cite{krizhevsky2012imagenet}   &  68.6  \\
CRC~\cite{zhang2011sparse} & 78.2   &    EPM~\cite{sharma2013expanded}   & 72.3  \\
CROC~\cite{chi2014classification}  &   79.2    &   ASPD~\cite{khan2015recognizing}   & 75.4  \\
ProCRC~\cite{cai2016probabilistic}  &   80.9    &   VGG19~\cite{simonyan2015very}   & 77.2  \\
SA-CRC~\cite{akhtar2017efficient}  &    77.3   & ANCR     &  \textbf{81.3} \\
\hline
\end{tabular}
\end{table}

\subsection{Experiments on the CUB-200-2011 Dataset}
\begin{figure}[!t]
\centering
\includegraphics[width=3in]{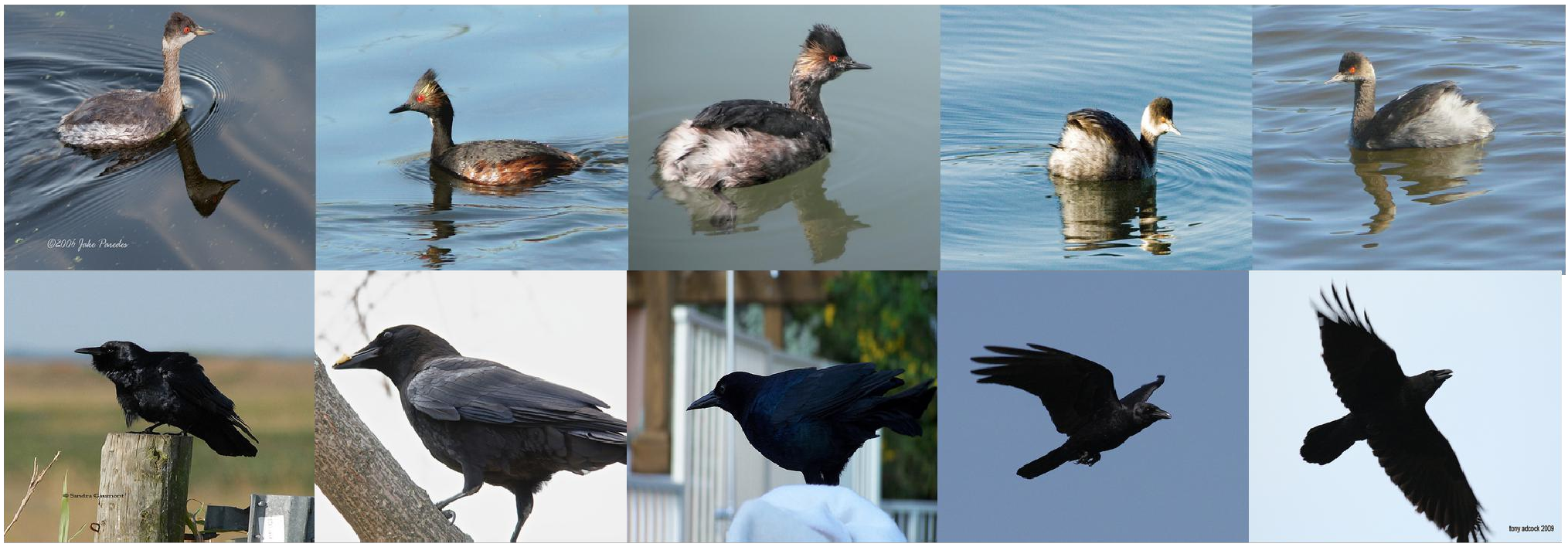}
\caption{Example images from the CUB-200-2011 dataset.}
\label{fig:exam_Cub}
\end{figure}

The Caltech-UCSD Birds 200 (CUB-200-2011) ~\cite{WahCUB_200_2011} is a representative fine-grained object classification dataset with images of 200 bird categories, there are 11,788 images and each category has about 60 images. This dataset is an extended version of the CUB-200 dataset. Some example images of the dataset are shown in Fig.~\ref{fig:exam_Cub}. We use the publicly available split~\cite{WahCUB_200_2011}, which uses nearly half of the images in this dataset as the training samples and the other half as the test samples. Features are extracted by using the pre-trained VGG-19 network and the balancing parameter $\lambda$ of ANCR is set to 0.001. POOF~\cite{berg2013poof}, FV-CNN~\cite{cimpoi2015deep}, VGG-19~\cite{simonyan2015very} and PN-CNN~\cite{branson2014bird} are also included for comparison. Experimental results are shown in Table~\ref{tab:tab_cub200}. According to the table, we can see that ANCR achieves the best recognition accuracy on the CUB-200-2011 dataset.

\begin{table}[!t]
\renewcommand{\arraystretch}{1.3}
\caption{A comparison of the proposed method in terms of recognition accuracy (\%) on the CUB-200-2011 dataset.}
\label{tab:tab_cub200}
\centering
\begin{tabular}{lc|cc}
\hline
Methods     & Accuracy & Methods        & Accuracy        \\ \hline
Softmax  &   72.1  &  CCRC~\cite{yuan2018a}  & 75.4   \\
NSC~\cite{lee2005acquiring}  &   74.5  &  NRC~\cite{xu2019sparse}   &  78.3  \\
SRC~\cite{wright2009robust}  &  76.0  &   POOF~\cite{berg2013poof}  & 56.9   \\
CRC~\cite{zhang2011sparse} & 76.2   &  FV-CNN~\cite{cimpoi2015deep}      &  66.7 \\
CROC~\cite{chi2014classification}  &   76.2    &  VGG19~\cite{simonyan2015very}     & 71.9  \\
ProCRC~\cite{cai2016probabilistic}  &  78.3     &  PN-CNN~\cite{branson2014bird}     &  75.7 \\
SA-CRC~\cite{akhtar2017efficient}  &    75.5   & ANCR     &  \textbf{78.6} \\
\hline
\end{tabular}
\end{table}

\subsection{Experiments on the Aircraft Dataset}
\begin{figure}[!h]
\centering
\includegraphics[width=3in]{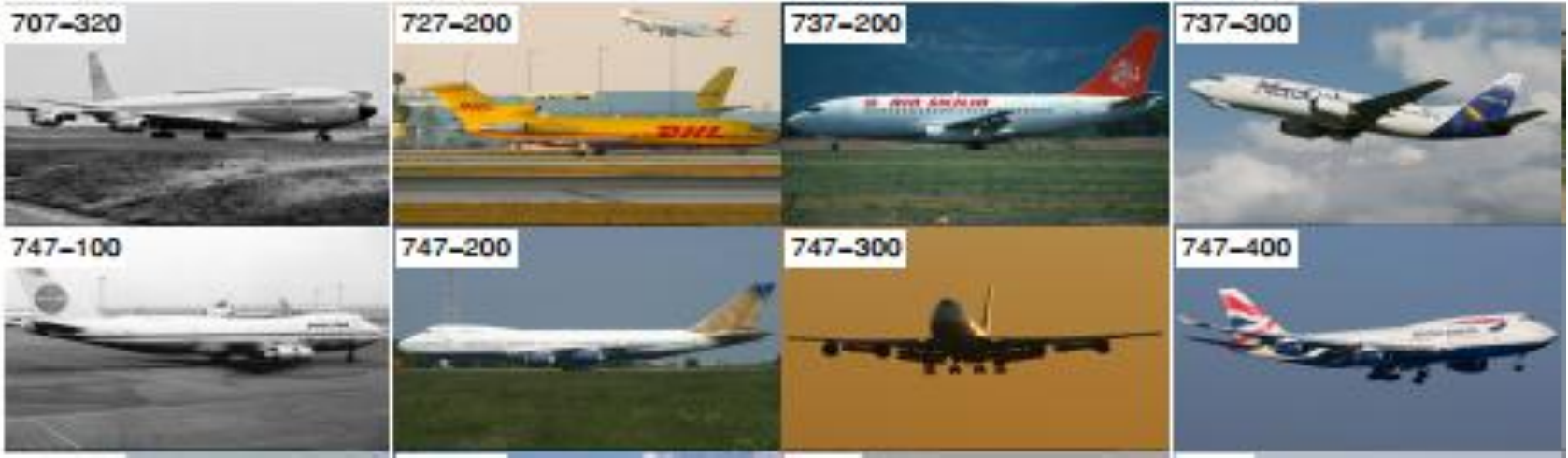}
\caption{Example images from Aircraft dataset.}
\label{fig:exam_Air}
\end{figure}

The Aircraft dataset~\cite{maji2013fine} contains 10,000 images of 100 different aircraft model variants, example images of this dataset are shown in Fig.~\ref{fig:exam_Air}. Following the experimental settings in~\cite{xu2019sparse}, 6667 images are used for training and 3333 images for testing. The image features are extracted by the pre-trained VGG-16 network. The balancing parameter $\lambda$ of ANCR is set to 0.001. We also compare our method with VGG-16~\cite{simonyan2015very}, Symbiotic~\cite{chai2013symbiotic}, FV-FGC~\cite{gosselin2014revisiting} and B-CNN~\cite{lin2015bilinear}. The recognition accuracy of each approach is presented in Table~\ref{tab:tab_aircraft}. Again, our proposed ANCR method outperforms all the other approaches.

\begin{table}[!t]
\renewcommand{\arraystretch}{1.3}
\caption{A comparison of the proposed method in terms of recognition accuracy (\%) on the Aircraft dataset.}
\label{tab:tab_aircraft}
\centering
\begin{tabular}{lc|cc}
\hline
Methods     & Accuracy & Methods        & Accuracy        \\ \hline
Softmax  &   85.6  &  CCRC~\cite{yuan2018a}  &  87.5  \\
NSC~\cite{lee2005acquiring}  & 85.5    &  NRC~\cite{xu2019sparse}   &  87.3  \\
SRC~\cite{wright2009robust}  &  86.1  &   VGG16~\cite{simonyan2015very}  &  85.6  \\
CRC~\cite{zhang2011sparse} &  86.7  &   Symbiotic~\cite{chai2013symbiotic}    & 72.5  \\
CROC~\cite{chi2014classification}  &    86.9   &  FV-FGC~\cite{gosselin2014revisiting}    & 80.7  \\
ProCRC~\cite{cai2016probabilistic}  &  86.8   & B-CNN~\cite{lin2015bilinear} &  84.1  \\
SA-CRC~\cite{akhtar2017efficient}  &  86.9     & ANCR     &  \textbf{87.7} \\
\hline
\end{tabular}
\end{table}

\subsection{Experiments on the Cars Dataset}
\begin{figure}[!h]
\centering
\includegraphics[width=3in]{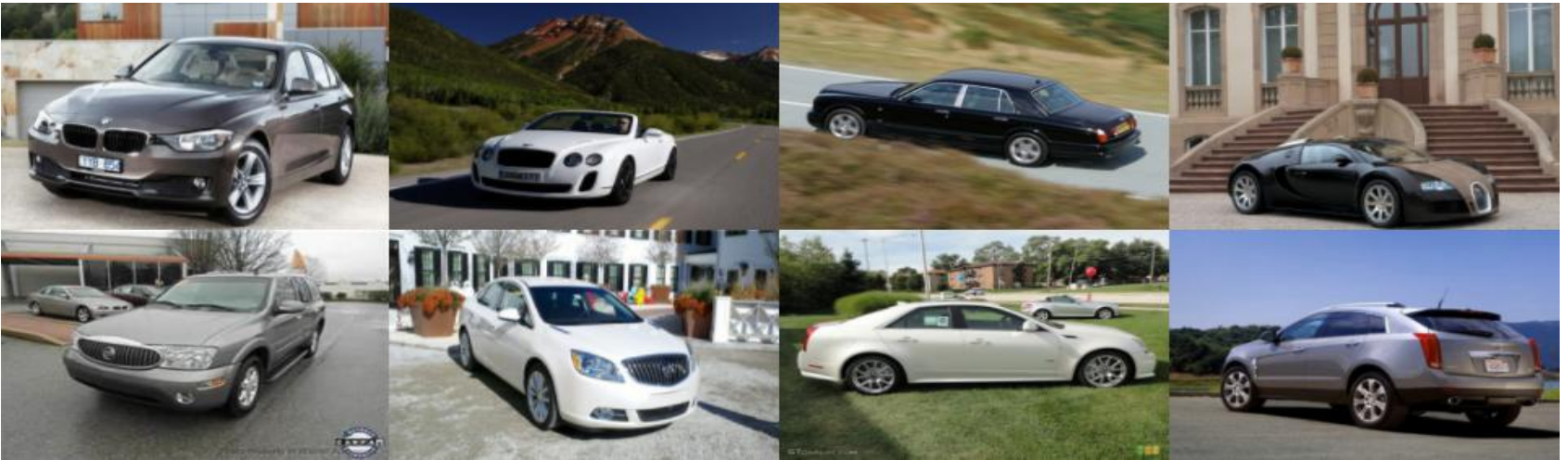}
\caption{Example images from the Cars dataset.}
\label{fig:exam_Cars}
\end{figure}

The Cars dataset~\cite{krause20133d} contains 16,185 images of 196 classes of cars, some example images of this dataset are shown in Fig.~\ref{fig:exam_Cars}. Following the standard evaluation protocol~\cite{krause20133d}, 8144 images are used for training and the other 8041 images are used for testing. Image features are extracted via a pre-trained VGG-16 network. The balancing parameter $\lambda$ of ANCR is set to 0.001. As on the Aircraft dataset, we compare the proposed method with VGG-16~\cite{simonyan2015very}, Symbiotic~\cite{chai2013symbiotic}, FV-FGC~\cite{gosselin2014revisiting} and B-CNN~\cite{lin2015bilinear} on this dataset. The results are reported in terms of accuracy in Table~\ref{tab:tab_cars}.  It can be seen that our proposed ANCR method achieves comparable performance as NRC, and it is superior to the other methods.

\begin{table}[!t]
\renewcommand{\arraystretch}{1.3}
\caption{A comparison of the proposed method in terms of recognition accuracy (\%) on the Cars dataset.}
\label{tab:tab_cars}
\centering
\begin{tabular}{lc|cc}
\hline
Methods     & Accuracy & Methods        & Accuracy        \\ \hline
Softmax  &  88.7   &  CCRC~\cite{yuan2018a}  &  90.6  \\
NSC~\cite{lee2005acquiring}  &   88.3  &  NRC~\cite{xu2019sparse}   & \textbf{90.7}   \\
SRC~\cite{wright2009robust}  &  89.2  &  VGG16~\cite{simonyan2015very}   &  88.7  \\
CRC~\cite{zhang2011sparse} &  90.0  &   Symbiotic~\cite{chai2013symbiotic}    &  78.0 \\
CROC~\cite{chi2014classification}  &   90.3    &  FV-FGC~\cite{gosselin2014revisiting}       & 82.7  \\
ProCRC~\cite{cai2016probabilistic}  &  90.1     &   B-CNN~\cite{lin2015bilinear}   &  90.6  \\
SA-CRC~\cite{akhtar2017efficient}  &   90.4   & ANCR     &  \textbf{90.7} \\
\hline
\end{tabular}
\end{table}

\subsection{Ablation Study}
In this subsection, we demonstrate the effectiveness of the non-negative constraint and affine constraint in ANCR by conducting ablation studies. To explore the impact of each constraint, we compare ANCR with several baseline models.

By discarding both the non-negative and affine constraints in Eq.~(\ref{obj_ancr}), we can obtain the first baseline model,
\begin{equation}
\label{obj_crc}
\underset{\boldsymbol{c}}{\textrm{min}} \ \left \| \boldsymbol{y}-\mathbf{X}\boldsymbol{c} \right \|_2^2+\lambda\left \| \boldsymbol{c} \right \|_2^2,
\end{equation}
which actually is the objective function of CRC, and Eq.~(\ref{obj_crc}) has the following closed-form solution,
\begin{equation}
\label{eq:solu_crc}
\boldsymbol{c}=(\mathbf{X}^T\mathbf{X}+\lambda \mathbf{I})^{-1}\mathbf{X}^T\boldsymbol{y}.
\end{equation}

When removing the non-negative constraint from Eq.~(\ref{obj_ancr}), we obtain the following problem,
\begin{equation}
\label{obj_acr}
\underset{\boldsymbol{c}}{\textrm{min}} \ \left \| \boldsymbol{y}-\mathbf{X}\boldsymbol{c} \right \|_2^2+\lambda\left \| \boldsymbol{c} \right \|_2^2, \ \textrm{s.t.} \ \boldsymbol{1}^T\boldsymbol{c}=1,
\end{equation}
which can be called the affine collaborative representation (ACR) model. By algebraic manipulations, one can derive the following closed-form solution to ACR,
\begin{equation}
\label{solu_acr}
\boldsymbol{c}=\hat{\boldsymbol{c}}/\mathbf{1}^T\hat{\boldsymbol{c}}, \ \hat{\boldsymbol{c}}=(\mathbf{M}+\lambda \mathbf{I})^{-1}\mathbf{1},
\end{equation}
where $\mathbf{M}=(\mathbf{X}^T-\mathbf{1}\boldsymbol{y}^T)(\mathbf{X}^T-\mathbf{1}\boldsymbol{y}^T)^T$.

By omitting the affine constraint in Eq.~(\ref{obj_ancr}), we get the following problem,
\begin{equation}
\label{obj_ncr}
\underset{\boldsymbol{c}}{\textrm{min}} \ \left \| \boldsymbol{y}-\mathbf{X}\boldsymbol{c} \right \|_2^2+\lambda\left \| \boldsymbol{c} \right \|_2^2, \ \textrm{s.t.} \ \boldsymbol{c}\geq 0,
\end{equation}
which is termed as the non-negative collaborative representation (NCR) model. Similar to the proposed ANCR method, NCR can be solved by the ADMM algorithm~\cite{boyd2011distributed}.

ANCR and the above three baseline models are summarized in Table~\ref{tab:tab_models}. Experiments are conducted on the six datasets and the experimental results are reported in Tables~\ref{tab:ab_ar}-\ref{tab:ab_four}, respectively. Based on the experimental results, the following observations can be made.

(1) NCR outperforms CRC in most cases, which indicates that the coding vector obtained by NCR contains more discriminative information than that of CRC, thus enhancing the classification capability of CRC.

(2) ACR is inferior to CRC except on the Aircraft and Cars datasets, which shows that by introducing just the affine constraint into the framework of CRC, the solution cannot guarantee that the classification accuracy be consistently improved.

(3) Our proposed ANCR method performs better in recognition accuracy than the other three baseline models. This demonstrates the role of both, the non-negative regularization and the affine constraint in our ANCR method.

\begin{table}[!t]
\renewcommand{\arraystretch}{1.3}
\caption{Summary of our ANCR and the three baseline models CRC, ACR and NCR.}
\label{tab:tab_models}
\centering
\begin{tabular}{l|l|ll}
\hline
\multirow{2}{*}{Models} & \multirow{2}{*}{Regularization terms} & \multicolumn{2}{c}{Constraints}            \\
                    &                        & \multicolumn{1}{c|}{$\boldsymbol{c} \geq 0$} & $\boldsymbol{1}^T\boldsymbol{c}=1$ \\ \hline
CRC~\cite{zhang2011sparse}                  & \multirow{4}{*}{$\left \| \boldsymbol{y}-\mathbf{X}\boldsymbol{c} \right \|_2^2+\lambda\left \| \boldsymbol{c} \right \|_2^2$}      & \multicolumn{1}{c|}{\xmark} &    \xmark        \\
ACR                 &                        & \multicolumn{1}{c|}{\xmark} &    \cmark        \\
NCR                &                        & \multicolumn{1}{c|}{\cmark} &     \xmark       \\
ANCR               &                        & \multicolumn{1}{c|}{\cmark} &     \cmark       \\
\hline
\end{tabular}
\end{table}

\begin{table}[!t]
\renewcommand{\arraystretch}{1.3}
\caption{Recognition accuracy (\%) of CRC, ACR, NCR and ANCR on the AR database.}
\label{tab:ab_ar}
\centering
\begin{tabular}{lccc}
\hline
Dimensions     & 54  & 120        & 300        \\ \hline
CRC~\cite{zhang2011sparse}  &  80.3   & 90.1   &  93.8  \\
ACR  &   78.5   &  88.5    & 91.4  \\
NCR &   85.2  &   \textbf{91.3}   &  93.3 \\
ANCR &  \textbf{86.0}   &  \textbf{91.3}  &   \textbf{94.0}  \\
\hline
\end{tabular}
\end{table}

\begin{table}[!t]
\renewcommand{\arraystretch}{1.3}
\caption{Recognition accuracy (\%) of CRC, ACR, NCR and ANCR on the USPS dataset.}
\label{tab:ab_usps}
\centering
\begin{tabular}{lcccc}
\hline
$N$     & 50  & 100    & 200     & 300        \\ \hline
CRC~\cite{zhang2011sparse}   &   89.8   &  90.8  &  91.5 &  91.5 \\
ACR  &    85.6     &  87.9    & 89.3  & 89.5  \\
NCR   & 90.2   &   91.6   & 92.6   & 92.9   \\
ANCR  &  \textbf{92.1}   & \textbf{93.0}   & \textbf{93.5}  & \textbf{94.3}  \\
\hline
\end{tabular}
\end{table}

\begin{table}[!t]
\renewcommand{\arraystretch}{1.3}
\caption{Recognition accuracy (\%) of CRC, ACR, NCR and ANCR on the four large scale datasets.}
\label{tab:ab_four}
\centering
\begin{tabular}{lcccc}
\hline
Methods    &    Stanford 40   & CUB-200-2011    &   Aircraft  &  Cars \\ \hline
CRC~\cite{zhang2011sparse}  &    78.2    &  76.2   &  86.7    &  90.0   \\
ACR   &    66.8    &  70.4    &   87.4  & 90.5 \\
NCR  &   81.1    &   78.4   &  87.3  &  90.6  \\
ANCR   &  \textbf{81.3}   &  \textbf{78.6}  & \textbf{87.7}  & \textbf{90.7}  \\
\hline
\end{tabular}
\end{table}

\subsection{Convergence and Parameter Sensitiveness Analysis}
\begin{figure}[!t]
\centering
\includegraphics[width=3.5in]{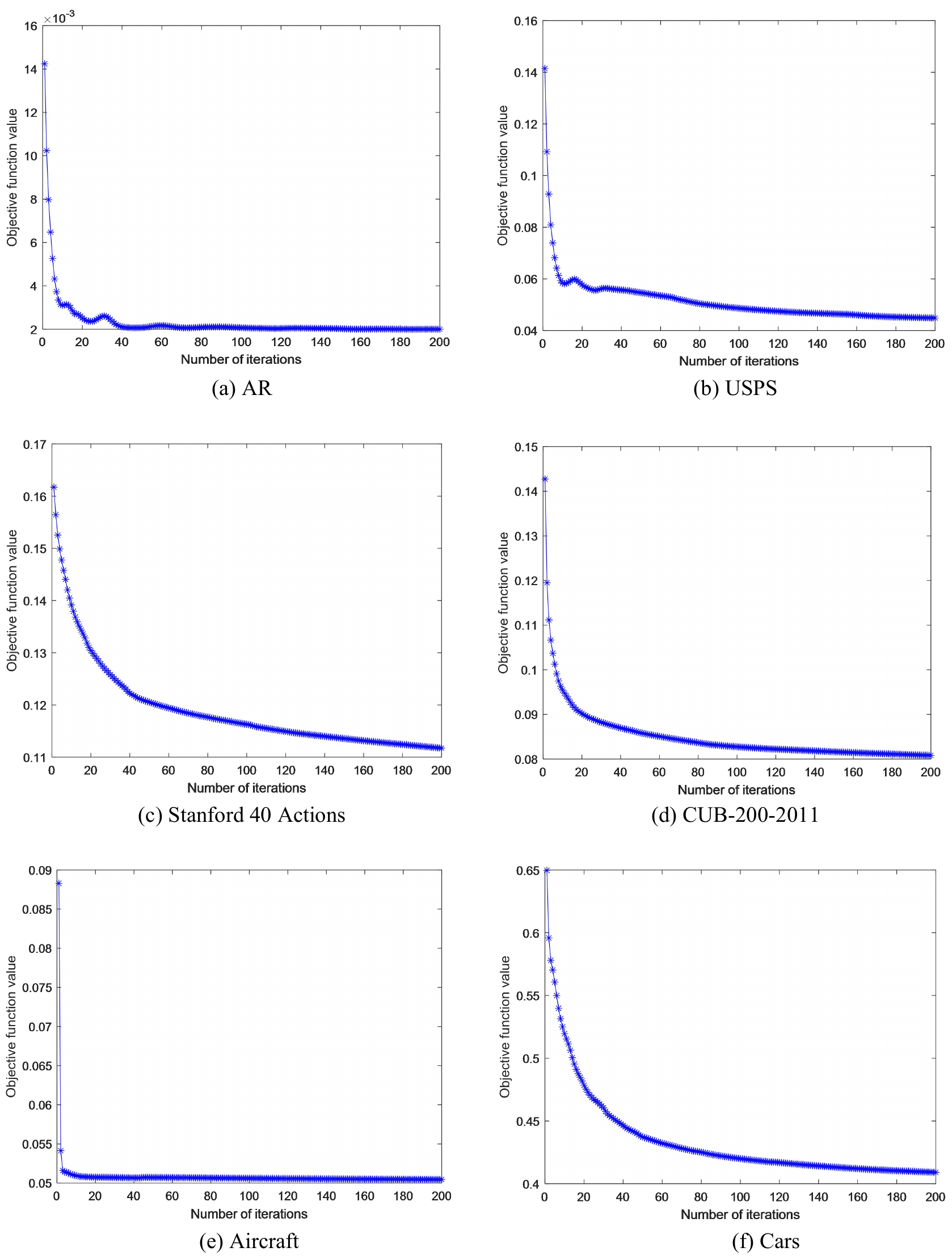}
\caption{Convergence curves of ANCR on different datasets: (a)-(f) are the convergence curves obtained in the experiments performed on the AR, USPS, Stanford 40 Actions, CUB-200-2011, Aircraft and Cars datasets, respectively.}
\label{fig:converg}
\end{figure}

The convergence of ADMM with two variable has been proven by Boyd \etal~\cite{boyd2011distributed}. Here we present the convergence curves on the six datasets in Fig.~\ref{fig:converg}. We can see that the objective function value of ANCR gradually decreases with the increasing number of iterations, which empirically demonstrates the convergence of the proposed ANCR approach.

\begin{figure}[!t]
\centering
\includegraphics[width=3.5in]{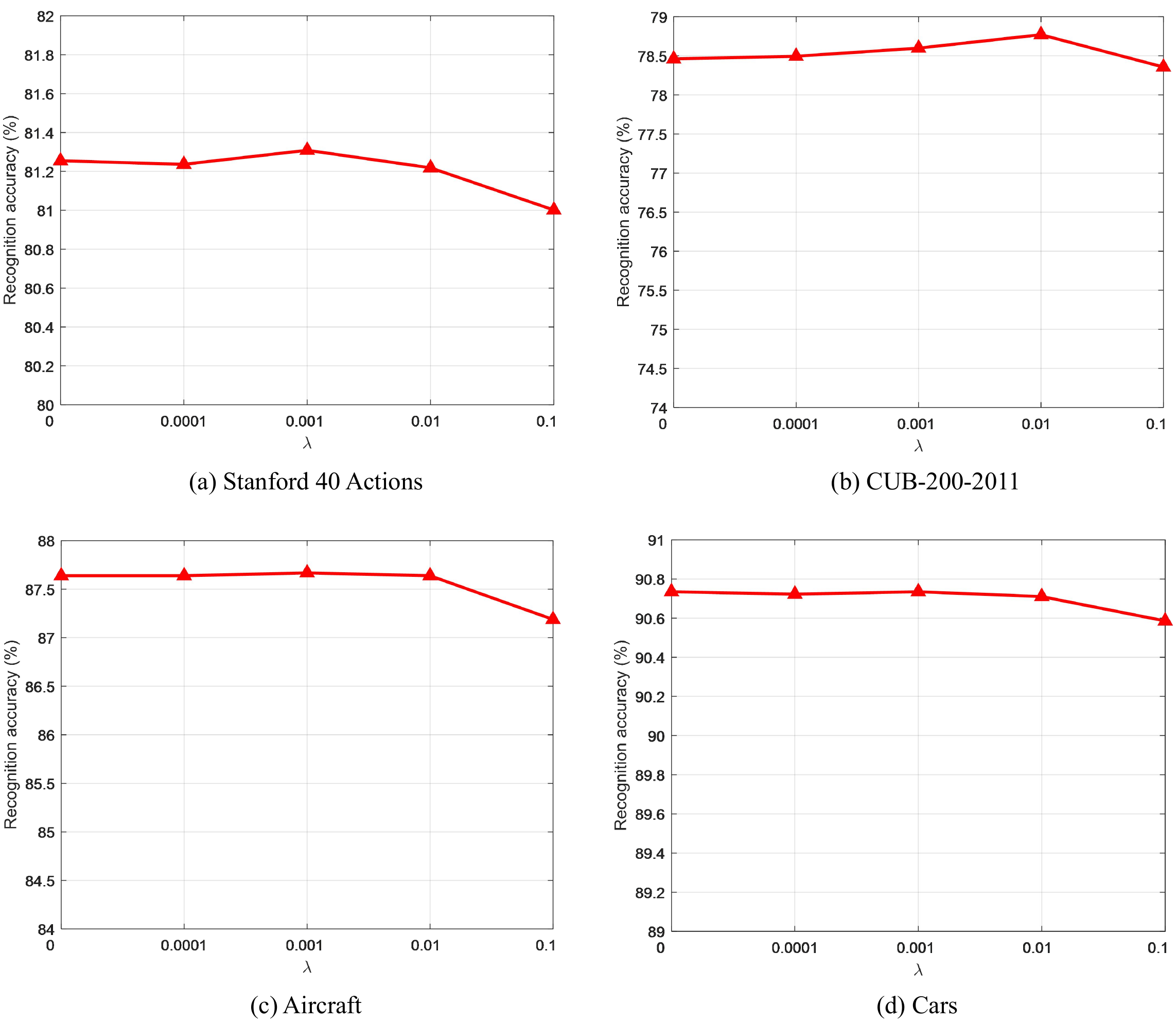}
\caption{Recognition accuracy (\%) of ANCR with varying parameter $\lambda$ on (a) Stanford 40 Actions (b) CUB-200-2011 (c) Aircraft and (d) Cars datasets, respectively.}
\label{fig:param}
\end{figure}

To examine how the balancing parameter $\lambda$ influences the performance of ANCR, we conducted experiments on the four large-scale datasets. The experimental settings are the same as those described  in the above sub-sections. Fig.~\ref{fig:param} plots the recognition accuracy with varying $\lambda$. We can see that ANCR returns stable results for a wide range of $\lambda$ values, \ie, [0.0001,0.01]. When $\lambda$ increases from 0.01 to 0.1, the performance of ANCR drops a little. Larger value of $\lambda$ means that ANCR will emphasize the $\ell_2$-norm of the coding vector, which could undermine the collaborative mechanism of all the training samples in representing a test sample. Therefore, we set a relatively small value for $\lambda$ in our experiments.

\section{Conclusion}
\label{Sec_5}
To further enhance the classification performance of NRC, in this paper, we developed an affine non-negative collaborative representation (ANCR) model. ANCR is derived by introducing a regularizer on the coding vector, as well as an affine constraint into the formulation of NRC. Our proposed ANCR method is solved elegantly by ADMM. The experimental results on six well-known datasets demonstrate the superiority of the proposed ANCR method over NRC and traditional representation based classification methods. The proposed method also outperforms some deep learning based approaches.


\section*{Acknowledgments}
The work was supported in part by the National Natural Science Foundation of China (61672265, U1836218, 61902153), in part by the 111 Project of the Ministry of Education of China (B12018), in part by the EPSRC programme grant (EP/N007743/1) and in part by the EPSRC/dstl/MURI project (EP/R018456/1).

{\small
\bibliographystyle{IEEEtran}
\bibliography{mybib}
}



%



\end{document}